\pdfoutput=1
\documentclass[11pt,a4paper]{article}
\usepackage{times,latexsym}
\usepackage{url}
\usepackage{import}
\usepackage{fancyvrb}
\usepackage[T1]{fontenc}
\usepackage{amssymb}
\usepackage{comment}
\usepackage{textcomp}

\usepackage[acceptedWithA]{tacl2018v2}

\usepackage{xspace,mfirstuc,tabulary}

\newif\iftaclinstructions
\taclinstructionsfalse %
\iftaclinstructions
\renewcommand{\confidential}{}
\renewcommand{\anonsubtext}{(No author info supplied here, for consistency with
TACL-submission anonymization requirements)}
\newcommand{\instr}
\fi

\iftaclpubformat %

\else

\fi

\newcommand{\ppceme}{PPCEME\xspace}
\newcommand{\ppcmbe}{PPCMBE\xspace}
\newcommand{\ppcme}{PPCME2\xspace}
\newcommand{\ptb}{PTB\xspace}
\newcommand{\eebo}{EEBO\xspace}
\newcommand{\corpussearch}{CorpusSearch\xspace}
\newcommand{\pos}{POS\xspace}
\newcommand{\evalb}{evalb\xspace}
\newcommand{\elmo}{ELMo\xspace}
\newcommand{\allennlp}{Allennlp\xspace}

\newcommand{\vtick}{\textquotesingle}

\title{Parsing Early Modern English for Linguistic Search}

\author{
 Seth Kulick and Neville Ryant \\ \\
 Linguistic Data Consortium, University of Pennsylvania \\
  {\sf \{skulick,nryant\}@ldc.upenn.edu} \\
}
\date{}

\begin{document}
\maketitle

\begin{abstract}
 We investigate the question of whether advances in NLP over the last few years make it possible to vastly increase the size of data usable for research in historical syntax. 
 This brings together many of the usual tools in NLP - word embeddings, part-of-speech tagging, and parsing - in the service of linguistic queries over automatically annotated corpora.  We train a \pos tagger and parser on a corpus of historical English, using \elmo embeddings trained over a billion words of similar text.  The evaluation is based on the standard metrics, as well as on the accuracy of the query searches using the
 parsed data.  
\end{abstract}

\section{Introduction}
\label{sec:intro}

Historical corpora manually annotated for syntactic information, such as the
Penn Parsed Corpus of Early Modern English (\ppceme) \cite{ppceme}, 
are an important resource for research in language change.  However, the total size of all such annotated corpora amounts to only a few million words, which represents a paltry portion of the billion words or so of Early Modern English text that is available. Moreover, syntactic annotation is particularly expensive and time consuming to produce.   Annotating a billion words with the same procedure used for the 
annotation of \ppceme would take about a millennium.   In this work we address the question of whether  current NLP tools can be used to instead automatically annotate very large amounts of additional material with  high enough accuracy that the linguistic searches of interest can be done on the automatically annotated corpora.

The work reported here is based on two corpora.  The first is the \ppceme, which  consists of about 1.9 million words of text, covering the time period from 1501-1719,  manually annotated for phrase structure.  The annotation principles are similar to that of the Penn Treebank (\ptb) \cite{marcus-etal-1993-building}, but also with various differences in part-of-speech (\pos) and syntactic tree annotation, resulting from the corpus being designed for linguistic research. Researchers in the line of work initiated by \citep{kroch89} utilize this corpus by searching the trees for various kinds of syntactic constructions, which can reveal information about changes over the time period of the corpus.  This search is done with \corpussearch \cite{corpussearch}, which allows researchers to search any \ptb-style corpus using boolean combinations of the standard syntactic relations
of dominance and precedence.

The other corpus involved in this work is the Early English Books Online (\eebo) \cite{eebo}, consisting of about 1.1 billion words of text from the time period from 1475-1700. EEBO is clearly a huge a potential source of new data for linguistic
research.  At the moment, however, its potential remains unrealized because 
it is not annotated for syntactic structure.\footnote{Though available in at least some form to members of the Text Creation Partnership since the early 2000s and to the general public since 2009, out of 600+ search results on Google scholar, only a handful (e.g., \citet{ecay}) involve the application of NLP technologies, usually \pos
tagging. \label{fn:utilization}}

Our goal is therefore to create structures for the \eebo corpus in the same style as the \ppceme annotation, which can therefore be used as input for \corpussearch queries.   While other approaches are
possible\footnote{E.g., converting the treebanks and \corpussearch queries to a  dependency representation.}, we take the straightforward approach here of using the \ppceme as training material for a \pos tagger and parser.  These are interesting test cases for NLP tools as both the annotation principles and data differ significantly from contemporary English; e.g., much greater spelling variation.

We approach evaluation of the \pos tagger and parser models from two complementary perspectives. First, we adopt the traditional approach of defining a training/dev/test split for the corpus (in this case, \ppceme), training on the training section, and evaluating the resulting models on the dev and test sections using the conventional metrics - accuracy for \pos tagging and evalb \cite{evalb} for the parsing.  We also extend the parsing analysis to the parser's ability to recover the function tags in the syntactic trees (indicating such 
information as subject, object, question, imperative, etc.) While there has been some work on recovering function tags in the parser output \cite{blaheta-charniak-2000-assigning,merlo-musillo-2005-accurate,gabbard-etal-2006-fully} they are typically left out of the evaluation.  However, some of the 
\corpussearch queries refer to the function tags of  nonterminals,
and so it is essential that the parser be able to automatically supply at least some of the most critical ones. 

In addition to the conventional approach, we evaluate the models using performance on a downstream task that is of interest to historical syntacticians: detecting syntactic structures of particular interest using \corpussearch.  Syntacticians identified a set of queries of particular interest, which were then run on the dev and test sections of \ppceme using both the gold syntactic trees and the trees from the \pos-tagged and parsed version of the corpus. We then compute precision, recall, and F1 for these queries, which give us a more meaningful measure of the quality of the tagger and parser output for \eebo.

\label{sec:corpus}
\section{Corpus Preparation}
In this section we describe the the main modifications made to the \ppceme and \eebo corpora for this work. We refer to appendix \ref{app:corpus} for some of the details, and we will also be releasing the code that was used for the corpora modifications.

\subsection{\ppceme}
\label{sec:prep:ppceme}
Here we describe the slight modifications we made to the \ppceme corpus, the corpus split we used for further work, and the tokenization developed based on the \ppceme, for use with \eebo. 

\subsubsection{Corpus Transformations}
\label{sec:prep:transformations}
The \ppceme is one of several corpora designed for historical linguistics research  that share annotation styles and design decisions.  Others include the
second edition of the Penn-Helsinki Parsed Corpus of Middle English \cite{ppcme2} (\ppcme) and the Penn Parsed Corpus of Modern British English \cite{ppcmbe} (\ppcmbe).  Previous work with \ppcme and \ppcmbe  has discussed characteristics of these corpora that differ from that of the \ptb and the impact it might have on models trained on these corpora \cite{moon-baldridge-2007-part,kulick-etal-2014-penn,yang-eisenstein-2016-part}.  A focus of this earlier work has been on how to transform the annotation to be closer to that of the \ptb, such as by transforming the phrase structure \cite{kulick-etal-2014-penn} or by mapping the \ppcme \pos tag set into that of \ptb \cite{moon-baldridge-2007-part,yang-eisenstein-2016-part}. 

In this work we approach this problem from a different angle. Since the output of our parser is used as input for the \corpussearch queries, we aimed to
make as few changes as possible to the source material for training the 
models. However, as the \ppceme \pos tag set is much larger than that usually used in parsing work on  \ptb (353 vs 36), we did apply two sets of changes to reduce it from 353 tags to a more manageable 85 tags. The major cases of such complexity, along with our handling of them, are:

\paragraph{Complex Tags}
There are 210 complex tags, such as {\tt PRO+N} ({\tt hymself)},  {\tt WPRO+ADV+ADV} ({\tt whatsoever)}, and {\tt ADJ+NS} ({\tt gentlemen)}.\footnote{These tags reflect the changing nature of the orthography - e.g., in earlier time periods the latter might be spelled {\tt (ADJ gentle) (NS men)}.} However, while these tags are numerous, they cover only about 1\% of the corpus, and these complex tags do not enter into the linguistic searches of concern.  Therefore we simply replaced each complex tag with its rightmost component (e.g., changing the above to {\tt N, ADV, NS}).\footnote{\citet{yang-eisenstein-2016-part} converted each complex tag to its leftmost component.   We chose the rightmost component because compounds  follow the Righthand Head Rule, 
according to which the properties of a compound word depend on the compound's righthand child.}

\paragraph{Multiword sequences treated as unitary}
 There are also cases where words are standardly written as a single orthographic token and sometimes as multiple separate tokens. \ppceme represents the former case with a single \pos tag and the latter as a constituent whose non-terminal is the \pos tag, with the words given numbered segmented \pos tags  -- for example, {\tt (ADJ alone)} vs {\tt (ADJ (ADJ21 a) (ADJ22 lone))}. We modified all such tags by removing the numbers, and appending {\tt \_NT} to the  nonterminals, to clearly distinguish between labels used as \pos tags and those used as nonterminals.  In this example, the resulting structure would be  {\tt (ADJ\_NT (ADJ a) (ADJ lone))}. 

\paragraph{Greater tag specificity}
 The \ppceme tagset has much greater specificity for the verbal tags than the \ptb.  For example, in addition to the tag for the infinitive form of {\it do} ({\tt DO}), there are variations for present, past, imperative, present participle, passive participle, and perfect participle ({\tt DOP, DOD, DOI, DAG, DAN, DON}), and likewise for {\it be}, {\it have}, and verbs other than {\it be, do, have}.
  While for the previous two cases of tag complexity we modified the tags, in this case we do not.  The reason is that this tag specificity is used in the \corpussearch queries, and in Section \ref{sec:queries} we give some examples of how these tags are used.  Mapping the \pos tags to 
 \ptb tagset approximations as in \citet{yang-eisenstein-2016-part} would lose this information.
 
\subsubsection{Corpus Split}

\begin{table}[t]
    \centering
    {\small
    \begin{tabular}{|l|r|r|r|}
        \hline
        {\bf Section} &    {\bf \# Sents}  &  {\bf \# Tokens}  &      {\bf \% total} \\
        \hline
        Train         &            85,398  &        1,725,604  &          89.51\%  \\
        \hline
        Dev           &             5,474  &          100,324  &           5.20\%  \\
        \hline
        Test          &             4,864  &          101,867  &           5.28\%  \\
        \hline
        Total         &            95,736  &        1,927,795  &         100.00\%  \\
        \hline
    \end{tabular}}
    \caption{Train/dev/test partitioning of the \ppceme.}
    \label{tab:ppceme-split}
\end{table}

We split the transformed \ppceme into  training, dev, and test partitions with roughly the same percentages for each partition as in the standard \ptb   split\footnote{train (sections 2-21): 90.75\%, dev (section 22): 3.83\%, test (section 23): 5.41\% \label{fn:ptb-split}}; for sizes of each partition; see  Table \ref{tab:ppceme-split}. Texts were sampled from the full temporal extent of \ppcme, with the result that each partition includes texts from a variety of time periods. Note that this stands in contrast to the partitioning suggested by \citet{yang-eisenstein-2016-part}, who split the corpus into thirds by time period for the purposes of studying domain adaptation. Details of the procedure used to assign texts to the partitions are available in Appendix \ref{app:corpus:ppcme:part}. 

\subsubsection{Tokenization}
\label{sec:tokenizer}
In order for the \ppceme-trained \pos-tagger and parser to produce sensible output, we need \eebo data to be tokenized in a way consistent with \ppceme. As the size of \eebo renders manual or semiautomatic tokenization impractical, we developed a deterministic tokenizer\footnote{Future  work  could  consider  a  more  sophisticated  tokenizer,  perhaps  as part of a joint tokenization-POS tagging task.} that attempts to replicate the \ppceme tokenization guidelines insomuch as possible. As this tokenizer is based on the \ppceme, we discuss it in this section.

The tokenization scheme for \ppceme is in principle straightforward:
    \begin{enumerate}
        \item possessives are left attached (e.g, {\tt Queen{\vtick}s}) (unlike in the \ptb)
        \item punctuation is separated except in the case of abbreviations (e.g., {\tt Mr.}) or hyphens with larger tokens (e.g. {\tt Fitz-Morris}), or in some special cases (e.g., {\tt \&c})
        \item Roman numerals are kept as one token, although their use was quite different in this material than in modern texts, with decimal points inserted into the numbers or beginning and ending them, such as {\tt .xiiii.C.}
    \end{enumerate}
    
However, the non-standard nature of the material presents various difficulties. For instance, while a {\tt th{\vtick}} prefix usually indicates a tokenization for a variant of {\tt the} (e.g. {\tt th{\vtick}exchaung} is tokenized as {\tt th{\vtick} exchaung}), sometimes the apostrophe is missing (e.g., {\tt thafternoone} is tokenized as {\tt th afternoone}). However, a leading  {\tt th} is not always split off (e.g., {\tt thynkyth} remains as one token). There are other ambiguities as well, such as with 
{\tt its}, which is  is tokenized in the \ppceme 336 times as one token (the possessive pronoun) and 96 times as 
\verb+it s+ (pronoun and copula). In such ambiguous cases, we implement the most the common decision; e.g., always split off a {\tt th{\vtick}} prefix, while not doing so for {\tt th} without the apostrophe, and always leaving {\tt its}  as one token.  We also encoded the most common abbreviations so they would be kept as one token, and handled some special cases such as the Roman numerals.  

In the end we were able to replicate the \ppceme tokenization on 99.86\% of the words in the \ppceme.

\subsection{\eebo}
\label{sec:eebo}

In this section we describe the procedures used for extraction of text from the \eebo XML files, its segmentation and its tokenization and segmentation into sentences. This sentence segmented and tokenized version of \eebo is the version used as input to \elmo training and parsing.

\subsubsection{Text extraction, normalization, and tokenization}
\label{sec:eebo_extract_norm}
The \eebo XML files contain a great deal of metadata and markup in addition to the text. For each file, we extracted the core source information (title, author, date) and kept the text within the \verb+<P>+ tags, which gave at least a rough sense of what the document divisions were. We followed the procedure of an earlier approach to using \eebo \cite[p. 105-106]{ecay} in excluding some metadata and other material embedded in the text. 

We also followed \citet{ecay} in our handling of {\tt GAP} tags, which  are used to indicate the locations of OCR errors.  For example, the following XML:
    \begin{Verbatim}[fontsize=\small]
    Eccl
    <GAP DESC="illegible"  DISP="•"/>
    siasticall
    \end{Verbatim}
is transformed into:
    \begin{Verbatim}[fontsize=\small]
    Eccl•siasticall
    \end{Verbatim}
in which the OCR errors (gaps) are represented by the bullet character.

The extracted text then underwent unicode normalization to NFC form in order to eliminate spurious surface differences between tokens. The resulting text contained 642 unique characters, 381 of which occurred fewer than 200 times. Manual inspection of these low frequency characters revealed that while some of these made sense in context (e.g., within sections of Greek or Latin text), many seemed to be spurious characters due to OCR errors (e.g., {\sc white rectangle 0x25ad}). Consequently, we elected to filter out all sentences containing characters occurring fewer than 200 times.

We then tokenized the \eebo text using the tokenizer discussed in Section \ref{sec:tokenizer}. As the tokenizer was originally developed using \ppceme, manual inspected revealed a number of issues when used on the \eebo data. In particular, \eebo exhibits wider variation in Roman numerals, including ones ending in {\tt j} for standard {\tt i}, such as {\tt v.C.xlviij}. We modified the tokenizer to account for such cases.

\subsubsection{Sentence Segmentation}
\label{sec:sentseg}

\begin{table}[t]
    \centering
    {\small
    \begin{tabular}{|l|r|r|r|} \cline{3-4}
    \multicolumn{1}{l}{} &  \multicolumn{1}{r}{} & \multicolumn{2}{|c|}{Exclusion Reason} \\ \hline
      & Included & Char & Length \\ \hline
       Sents & 29,580,930 & 5,297 & 3,892 \\ \hline
      Words & 1,165,287,328 & 282,967 & 4,511,837 \\ \hline
    \end{tabular}}
    \caption{Statistics for \eebo, showing \# included and excluded due to either having a questionable character ($\leq$ 200 occurrences) or a sentence longer than 800 tokens.}
    \label{tab:eebo}
\end{table}

In order to render \eebo suitable for parsing and search, we implemented a rudimentary sentence segmentation by splitting on paragraph tags in the XML, then on all tokens consisting solely of a period, exclamation mark, or question mark. We also eliminated all sentences longer than 800 tokens as they tended to be pathological cases (e.g. a ``sentence'' consisting of a long list of map coordinates). As discussed in section \ref{sec:eebo_extract_norm}, we also filtered sentences containing characters occurring fewer than 200 times in \eebo. Table \ref{tab:eebo} shows the amount of data included and the relatively small amount excluded due to rare characters and sentence length.  

\section{\elmo Embeddings}
\label{sec:elmo}
In recent years, contextualized word embeddings \citep{peters2018deep,devlin2018bert} have driven significant improvements on downstream NLP tasks, including \pos tagging and parsing. Due to the significant overhead involved in training these representations, researchers often make use of pretrained models distributed by large companies, sometimes fine-tuned to the domain of interest. While this often produces perfectly satisfactory results, when the mismatch between the training (usually some combination of Wikipedia scraped web text) and test domains is large, significant improvements can be extracted by pretraining on the novel domain \citep{lee2019biobert,beltagy2019scibert,jin2019probing}. For this reason, we pretrained  \elmo embeddings using the same model configuration as \citet{peters2018deep} for 11 epochs\footnote{Corresponding to 2 weeks of training using four GTX 1080 GPUs.} using the entirety of \eebo. For downstream tasks, we use a linear interpolation of the outputs of the final two layers of the \elmo model with the interpolation weight learned during training for the task; the resultant embeddings have 1,024 dimensions.

\label{sec:pos_tagger}
\section{Part-of-speech Tagger}
Our POS tagger is based on the same LSTM-CRF architecture used for named entity tagging by \citet{peters2017semi} and \citet{peters2018deep}. Each token is represented as the concatenation of three types of embeddings: (1) word embeddings learned during training; (2) character level representations derived from a CNN operating on character embeddings \citep{ma2016end}; (3) \elmo embeddings. These representations are then fed through two bidrectional LSTM layers and then through a single linear layer that predicts a score for each \pos tag. Additionally, we incorporate parameters for the transitions between each pair of \pos tags so that the resulting model has the form of a linear chain conditional random field (CRF). Training is conducted using a sentence-level CRF loss computed using forward-backward, while Viterbi decoding is used for inference at test time.

\begin{table}[t]
    \centering
    {\small
    \begin{tabular}{|l|c|c|}  \hline
{\bf Embeddings}  &  {\bf Dev}    & {\bf Test}    \\ \hline
\eebo       & 98.10\%  & 97.95\% \\ \hline
original       & 96.97\%  & 96.81\% \\ \hline
none      & 96.05\%  & 95.92\% \\ \hline
    \end{tabular}}
    \caption{Per-tag accuracy for the \pos tagger on the \ppceme dev and test sections. The first column indicates the source of the \elmo embeddings: {\it \eebo}: \elmo pretrained on \eebo; {\it Original}: \elmo trained on 1B Word Benchmark; {\it None}: no \elmo embeddings.}
    \label{tab:pos}
\end{table}

The accuracy of the trained tagger on the \ppceme dev and test sets is presented in Table \ref{tab:pos}. As can be seen, the best score results from using the domain-specific \eebo embeddings, achieving an accuracy of 97.95\% on the \ppceme test set, which represents an improvement 1.14\%  absolute relative to \elmo embeddings trained on contemporary English and 2.03\% relative to using no \elmo embeddings at all. Examination of the confusion matrix on the test set for for tagger using \eebo embeddings revealed that most common tagging errors were confusion between {\tt N} and {\tt NPR} tags (common noun and proper noun), followed by {\tt N} and {\tt ADJ} and {\tt N} and {\tt FW} (foreign word). A similar pattern was observed for the tagger using the original embeddings, though the error rates were roughly double; additionally, the tagger using the original embeddings exhibited four times
as many instances of confusion between {\tt N} and {\tt VAG} (present participle).

\section{Parser}
\label{sec:parser}
For constituency parsing we use the reconciled span parser (RSP) architecture introduced in \citet{joshi2018extending}. Each token is represented by the concatenation of the \elmo embeddings from section \ref{sec:elmo} and 50-dimensional \pos embeddings, which are then fed into an encoder consisting of two bidirectional LSTM layers. Each possible span is then represented by the differences in the encoder states between the beginning and end of the span and a one-layer feedforward neural network used to predict the label, which may be either one of the non-terminals or $\varnothing$ (empty sequence). Span conflicts are resolved to produce a well-formed tree using a greedy algorithm as described in section 2.1 of \citet{joshi2018extending}.

As discussed in Section \ref{sec:intro}, it is necessary to retain  function tags in the parser output, for use 
with the \corpussearch queries. We therefore first discuss in more detail the  function tags in the \ppceme and how we integrated them into the parser model, before discussing the parsing results. 

\subsection{Function Tags}
\label{sec:parser:ftags}
While \ppceme has some function tags in common with the \ptb (e.g., {\tt SBJ}, {\tt LOC}, {\tt DIR}), it also has a  wider range of function tags, such as {\tt TMC} for ``tough movement complement'', {\tt EOP} for ``empty operator clause'', {\tt FRL} for ``free relative clause'', and so on.  Altogether there are 37 function tags in
\ppceme, compared to 20 in the \ptb.  In the \ppceme, parentheticals are annotated with a {\tt PRN} tag on their nonterminal, rather than having a {\tt PRN} node as in the \ptb.

For this first work with parsing the \ppceme, we focused on recovering just a subset of the function tags, which are listed in Table \ref{tab:ftags}. These are the tags required by the first set of \corpussearch queries that we are working with. We leave the problem of recovering the full range for future work.

\begin{table}[t]
    \centering
    {\small
    \begin{tabular}{|l|l|c|}
    \hline
    {\bf Tag}  &  {\bf Meaning}       & {\bf Used with}  \\
    \hline
    MAT        &  matrix clause       & IP, CP \\
    \hline
    SUB        &  subordinate clause  & IP \\
    \hline
    IMP        & imperative           & IP \\
    \hline
    INF        & infinitival          & IP \\
    \hline
    QUE        & question clause      & CP \\
    \hline
    SBJ        & subject              & NP \\
    \hline
    ACC        & accusative           & NP \\
    \hline
    DTV        & dative               & NP \\
    \hline 
    VOC        & vocative             & NP \\
    \hline
    PRN        & parenthetical        & IP, CP, NP, PP \\
    \hline
    \end{tabular}}
    \caption{Function tags retained for parsing}
    \label{tab:ftags}
\end{table}

Earlier work on recovering function tags  \cite{gabbard-etal-2006-fully}  used the simple approach of not deleting 
the function tags so that they were integrated into the model, in effect treating the labels with function tags as
atomic nonterminal labels (e.g.
{\tt NP-SBJ} as a label different than {\tt NP}).
Somewhat surprisingly, it made little difference in the \evalb accuracy of the parser
output compared to a model trained without the function tags.\footnote{Where the accuracy is evaluated by the \evalb implementation \cite{evalb},  which deletes all function tags for evaluation.} 

We follow this approach, and therefore modified the \allennlp parser to not delete any function tags in the training data, and modified the training data to include only the ten function tags.  

Finally, we note here that while we are including some of the function tags, we are following a long range of parsing work in not attempting to model the empty categories
and co-indexing in the treebank annotation. While this is not an issue for the current 
queries, it will likely be more so in future work, as discussed a bit in Section 
\ref{sec:queries:fullparses}.

\begin{table*}
\begin{minipage}[b]{.68\textwidth}
    \centering
    {\small
    \begin{tabular}{|l|c|c|c|c|c|c|}   \cline{2-7}
    \multicolumn{1}{c}{ } & 
    \multicolumn{3}{|c|}{{\bf With Function Tags}}  & 
    \multicolumn{3}{|c|} {{\bf Without Function Tags}} \\ \hline
        {\bf Embeddings}    &  {\bf Recall} &  {\bf Prec} &  {\bf F1} &  {\bf Recall} &  {\bf Prec} &  {\bf F1}  \\
        \hline
        (1) \eebo           &         89.45 &       90.37 &     89.91 &         89.88 &       89.85 &     89.86 \\
        \hline
        (2) original        &         86.74 &       87.65 &     87.19 &         86.77 &       87.56 &     87.17  \\
        \hline
        (3) none, tokens    &         83.27 &       87.07 &     85.13 &         83.56 &       86.04 &     84.78  \\
        \hline 
        (4) \eebo, tokens   &         89.50 &       90.01 &     89.75 &         89.67 &       90.09 &     89.88 \\
        \hline
        (5) Berkeley        &         82.80 &       83.15 &     82.97 &         82.90 &       83.80 &     83.35 \\
        \hline
    \end{tabular}}
    \caption{Evalb precision, recall, and F1 for the \ppceme dev section, with and without function tags. The first column indicates
            the source of the \elmo embeddings: (1): \elmo trained pretrained on \eebo; (2): \elmo trained on 1B Word Benchmark; 
            (3): no \elmo embeddings $+$ trained  embeddings and CNN embeddings; (4) \elmo trained on \eebo $+$  word embeddings and CNN embeddings; 
            (5) Berkeley parser trained on \eebo.}
    \label{tab:results:parsing}
    \end{minipage}
    \begin{minipage}[b]{.32\textwidth}
    \centering
    {\small 
    \begin{tabular}{|l|r|r|}  \hline
    {\bf Tag}  &   {\bf \# gold} &  {\bf F1} \\
    \hline
    SBJ        &            7907 &     98.63 \\
    \hline
    MAT        &            4817 &     98.82 \\
    \hline
    SUB        &            4663 &     98.99 \\
    \hline
    ACC        &            3963 &     96.59 \\
    \hline
    INF        &            1430 &     97.51 \\
    \hline
    PRN        &             693 &     87.86 \\
    \hline
    QUE        &             690 &     96.77 \\
    \hline
    IMP        &             539 &     98.12 \\
    \hline
    DTV        &             513 &     94.12 \\
    \hline
    VOC        &             352 &     97.58 \\
    \hline
    \end{tabular}}
    \caption{Function tag results for \ppceme dev set.}
    \label{tab:results:ftags}
    \end{minipage}
\end{table*}

\subsection{Results}
\label{sec:parser:results}

Results for the parser on the \ppceme dev section with and without the ten function tags are presented in Table \ref{tab:results:parsing}.  As was the case for \pos tagging, F1 is highest when using the \eebo trained \elmo embeddings (row (1)), achieving 89.91\% when using function tags, which represents an absolute improvement of 2.72\% relative to using \elmo trained on modern sources (row (2)). We also considered the effects on parsing performance of incorporating additional trained word embeddings and character CNN derived embeddings as was done for the \pos tagger. When the parser was trained using only these representation and no \elmo model at all (row (3)), the dropoff in performance is striking: nearly 4.8\% absolute. No improvement was noted when using the word embeddings and character CNN embeddings in conjunction with \elmo (row (4)), indicating that the contextual representations, when trained on in domain materials, are themselves sufficient for this task.

We also trained and evaluate the Berkeley parser \cite{petrov-klein-2007-improved},  as a way to gauge the improvements in parsing technology over the last decade.
As can be seen in row (5), the results of the Berkeley parser are quite a bit lower than even the worst performing of the neural variants.

As mentioned, we also trained the parser with the function tags removed from the training data, with the results shown in the rightside columns in  Table \ref{tab:results:parsing}.  Consistent with earlier results \cite{gabbard-etal-2006-fully}, the change in
F1 is fairly minimal, especially for the higher performing parsers.

The function tags results are shown in Table \ref{tab:results:ftags}.
These are scored using the same method as in \citet{gabbard-etal-2006-fully}, by comparing the function tags on the nonterminals for which the bare labels match in the gold and parsed trees.  For example, if a {\tt NP} bracket has the same span in the gold and parsed tree and  therefore considered  a match by evalb, the function tags on the gold and parsed versions are considered for the function tag evaluation. If, e.g., it the bracket label is {\tt NP-SBJ}  in both the gold and parsed versions it is a match, if it is {\tt NP-SBJ} in the gold but {\tt NP} in the parsed it is a recall error, and if it is the reverse then it is a precision error.  The \#gold column is the \# of occurrences in the \ppceme dev section gold trees.  Unsurprisingly, the {\tt PRN} tag is by far the worst-performing of the function tags.

\begin{figure*}
\begin{minipage}[c][16.5cm][t]{.45\textwidth}
\vspace*{\fill}
\scriptsize
{\bf finClause} \verb+IP-MAT*|IP-SUB*+ \\
{\bf subject}  \verb+NP-SBJ*+ \\
{\bf inf(initive)} \verb+BE|DO|HV|VB+ \\
{\bf finDo} \verb+DOD|DOP+ \\
{\bf finVerb} \verb+DOD|DOP|HVD|HVP|VBD|VBP+ \\
{\bf part(iciple)} \verb+DAN|HAN|VAN|BEN|DON|HVN|VBN+
\caption{Definitions for queries.  The last four are convenience groupings for \pos tags, for which tags ending in {\tt D} are the past form, ending in {\tt P} the present, ending in {\tt AN} the passive participle, and others ending in {\tt N} the passive participle.}
\label{fig:defns}
\begin{Verbatim}[fontsize=\scriptsize,commandchars=\\\{\}]
\fbox{\bf inverted}
{\bf (finClause ...finVerb...subject...)}
(CP-QUE-MAT 
   (IP-SUB (DOD did)
           (NEG not)
           (NP-SBJ (NPR Carpenter))
           (VB ask)
           (NP-DTV you)))

\fbox{\bf do-not}    
{\bf (finClause ...finDo NEG...inf|part...)}
(IP-SUB (NP-SBJ (PRO they))
        (DOP do)
        (NEG not)
        (VB perish))
        
\fbox{\bf verb-not}
{\bf (finClause ...finVerb NEG...),}
{\bf (no inf|part in finClause)}
(IP-MAT (NP-SBJ (PRO you))
        (DOP do)
        (NEG not)
        (NP-ACC (PRO$ your)
                (N dutie)))
        
(IP-MAT (NP-SBJ (PRO they))
        (VBP consider)
        (NEG not)
        (IP-INF (TO to)
                (VB cut)
                (NP-ACC 
                   (PRO it))))
\end{Verbatim}
\caption{Queries used for declarative clauses. These searches apply only to subtrees rooted in {\tt IP}.}
    \label{fig:queries1}
\end{minipage}
\hspace{0.5cm}
\begin{minipage}[c][16.5cm][t]{.45\textwidth}
\vspace*{\fill}
\begin{Verbatim}[fontsize=\scriptsize,commandchars=\\\{\}]
\fbox{\bf non-inverted}
{\bf (CP-QUE-MAT}
{\bf \ \ \ \  (IP-SUB ...subject ... finiteVerb)}
(no example tree)

\fbox{\bf do-subj}
{\bf (CP-QUE-MAT}
{\bf \ \ \ \  (IP-SUB ...finDo...subject... inf|part...))}
(CP-QUE-MAT
    (WNP-1 (WD What) (N Name))
    (IP-SUB 
        (DOD did)
        (NP-SBJ 
            (D the)
            (N Fellow)
            (PP (P with)
                (NP (D the)
                    (N Beard))))
        (VB tell)
        (NP-DTV (PRO thee))
        (CP-THT (C 0)
            (IP-SUB (NP-ACC *T*-1)
                    (NP-SBJ (PRO he))
                    (HVD had))))
    (. ?))   
   
\fbox{\bf verb-subj}
{\bf (CP-QUE-MAT}
{\bf \ \ \ \  (IP-SUB ...finVerb...subject ...))}
{\bf (no inf|part in IP-SUB)}
(CP-QUE-MAT
    (WADVP (WADV where))
    (IP-SUB (VBD came)
            (NP-SBJ (NPR Carpenter))
            (PP (P unto)
                (NP (PRO you))))
    (. ?))

\end{Verbatim}
\caption{Queries used for question clauses. These searches apply only to subtrees rooted in {\tt CP}.}
    \label{fig:queries2}
\end{minipage}
\end{figure*}

\begin{table*}
    \centering
    {\small
    \begin{tabular}{|r|r|r|r|r|r|r|r|r|}   
        \hline
                           & gold & non-gold &  \multicolumn{6}{|c|}{--------- gold vs. nongold ---------} \\ \hline
         search            & Hits & Hits & Match & Miss & FA & Recall & Prec  & F1 \\ \hline
         ignore-inverted   & 328 & 348  &  313  & 15  & 35 & 95.43 & 89.94  & 92.60 \\ \hline
         do-not   & 86  &  84  &   83  &   3 & 1  & 96.51 & 98.91  & 97.65 \\ \hline 
         verb-not & 88  &  89  &   84  &   4 & 5  & 95.45 & 94.38  & 94.92 \\ \hline
         do-subj  & 181 & 165  &  165  &  16 & 0  & 91.16 & 100.00 & 95.38 \\ \hline  
         verb-subject      &  55 &  52 &    50  &   5 & 2  & 90.91 & 96.15 & 93.46 \\ \hline
    \end{tabular}}
    \caption{Results comparing the searches run on the gold dev section (gold \pos tags, gold trees) with the dev section which has predicted \pos tags and parser generated trees.  The miss column is the number of hits found in the gold version that were not found in the non-gold version and the FA (false alarm) column is the number of hits found in the non-gold that were not in the gold.}
    \label{tab:results}
\end{table*}

\begin{figure*}
\begin{minipage}[b][10cm]{0.45\textwidth}
\vspace*{\fill}
\begin{Verbatim}[fontsize=\scriptsize,commandchars=\\\{\}]
    \fbox{GOLD}      \fbox{PARSED}
(CP-QUE-MAT     (IP-MAT ...
   (INTJ NO)       [rest is same]
   (, ,)                    
   (CONJ nor)               
   (IP-SUB
      (DOD did)        
      (NP-SBJ (Q no) (N body))
      (VB ask)
      (NP-DTV (PRO you))
      (IP-INF (TO to)
              (VB eat)))
   (. ?))
\end{Verbatim}    
\begin{Verbatim}[fontsize=\scriptsize,commandchars=\\\{\}]
  \fbox{GOLD}                 \fbox{PARSED}
(CP-QUE-MAT             (CP-QUE-MAT
   (WADVP-1                (ADVP
      (WADV Where))           (WADV Where))
   (IP-SUB
      (ADVP-LOC *T*-1)
      (DOP do)             (DOP do)
      (NP-SBJ (PRO you))   (NP-SBJ (PRO you))
      (VB live))           (VB live)
    (. ?))                 (. ?))
\end{Verbatim}
\caption{{\tt do-subj} recall errors}
\label{ds-recall}
\end{minipage}
\begin{minipage}[b][10cm]{0.45\textwidth}
\vspace*{\fill}
\begin{Verbatim}[fontsize=\scriptsize,commandchars=\\\{\}]
   \fbox{GOLD}                      \fbox{PARSED}
(CP-QUE-MAT               (CP-QUE-MAT
   (IP-SUB                   (IP-SUB
      (BEP Is)                  (BEP Is)
      (NP-SBJ-1                 (NP-SBJ
         (EX ther))                (EX ther))
      (NP-1                     (NP
         (QP                       (QP
            (ADVR to)                 (ADVR to)
            (Q muche))                (Q muche)))
         (CP-QUE-MAT-PRN        (VBP thynke)
            (IP-SUB-PRN         (NP-SBJ
               (VBP thynke)        (PRO you))
               (NP-SBJ          (PP (P for)
                  (PRO you))))     (NP
         (PP (P for)                  (D a)
            (NP                       (N kynge)))))
               (D a)
               (N kynge)))))
\end{Verbatim}    
\caption{{\tt verb-subject} recall error}
\label{vs-recall}
\end{minipage}
\end{figure*}

\section{Queries}
\label{sec:queries}
We cannot show here the full range of queries of interest to the linguists, and we instead focus on two groups of queries, having to do whether a verb moves to the position where {\tt do} might occur.  Figure \ref{fig:defns} shows some definitions used for convenience in the queries.  {\bf finClause} stands for nonterminals introducing a finite {\tt IP} clause, including {\tt IP-MAT} and {\tt IP-SUB} but excluding {\tt IP-INF}.  The last four definitions use subsets of the verbal \pos tags described in Section \ref{sec:prep:transformations}. 

\subsection{Declarative Clause queries}

Figure \ref{fig:queries1} shows the three queries used for classifying subtrees rooted in {\tt IP}. For each {\tt IP}, \corpussearch will carry out the queries in order, stopping if one is found.  That is, it will first check an {\tt IP} to see if it matches the {\tt inverted} query, and if not it will try the {\tt do-not} query, and if not, it will try the {\tt verb-not} query.  For each query, we show a somewhat simplified form of the query definition\footnote{We are  abstracting from the details of the \corpussearch specification, which is expressed in a series of boolean conditions.}, followed by one or (for {\tt verb-not}) two example trees that match the query.  {\tt inverted} is a ``helping'' query,
while {\tt do-not} and {\tt verb-not} are the two queries of interest.

{\tt inverted} finds cases of an inverted subject (coming after the verb), so  that they are removed from consideration for the other two queries.  {\tt IP}-rooted subtrees may have an inverted subject due to it being part of a question (with {\tt CP} as a parent),  as in the example for {\tt inverted}, with the {\bf subject} {\tt Carpenter} after the {\bf finVerb} {\tt did}.

\verb+do-not+ finds cases as in modern English, with an auxiliary {\tt do} before a \verb+NEG+ with
the {\tt infinitive} or {\tt participle} following. A simple example is shown, with the {\bf inf} {\tt perish}.  The example tree for {\tt inverted} would have been found by {\tt do-not} if it had  not already been classified as {\tt inverted}.

{\tt verb-not} finds cases in which the verb appears where the {\tt do} is in the {\tt do-not} case, with the finite verb before the {\tt NEG}. There is a negative condition specified in the \corpussearch query that there is no infinitive or participle in the clause.  In the first example, the {\bf NEG} follows the {\bf finVerb} {\tt do}, and in the second the {\bf NEG} 
follows the {\bf finVerb} {\tt consider}.

\subsection{Question queries}

Figure \ref{fig:queries2} shows the queries used for classifying subtrees rooted in {\tt CP}.  {\tt do-subj} and {\tt verb-subj} are  parallel to {\tt do-not} and {\tt verb-not} except that the positioning of the {\tt do} or the {\tt finiteVerb} is determined in relation to the subject rather than {\tt NEG}.   They are again mutually exlusive, in that \corpussearch will first consider whether a {\tt CP} matches {\tt non-inverted}, then whether it matches {\tt do-subj},
and then whether it matches {\tt verb-subj}.

{\tt do-subj} finds cases as in modern English, with an auxiliary {\tt do} before subject with the {\tt infinitive} or {\tt participle} following.  In the example shown the {\bf finDo} {\tt did} is before the {\bf subject} {\tt the Fellow with the Beard} which is before the {\bf inf} {\tt tell}

{\tt verb-subj} finds cases in which the verb appears where the {\tt do} is in the {\tt do-subj} case, with the finite verb before the subject, with a negative condition that there is no infinitive or participle in the clause.    In the example, the {\bf finVerb} {\tt came} precedes the {\bf subject} {\tt Carpenter}.

The analog of the  {\tt IP} {\tt inverted} query would be a {\tt non-inverted} query for the {\tt CP} nodes.  However, since this the question queries are on the {\tt CP} node, and the subject should always be inverted in a question, this does not occur in the gold corpus, and so we do not show an example here. However, it does come up as an error in the parse output, discussed in Section \ref{sec:queries:errors}.

\subsection{Query Results}
\label{sec:queries:results}
As described in the introduction, we evaluate the utility of this approach for the goal of linguistic research by running the \corpussearch queries on both the gold and parsed versions of the dev section.  The results of the queries on the gold section are the gold hits against which the hits found in the parsed version are compared, with separate scores for each query.  Table \ref{tab:results} shows the results for the dev section.  We leave out the results for the test section, which are very similar.

As the results show, the searches do quite well, although a more complete evaluation will involve 10-fold cross validation on the \ppceme instead of the one split we are using.  However, the current results show the potential of using  \corpussearch queries on the \eebo.  In the following section however we discuss some of the errors that occur and areas that need improvement in order to reach this goal.

\subsection{Query Search Errors}
\label{sec:queries:errors}
The query search errors are of course the result of the difference between the gold version of the trees and the non-gold - i.e., resulting from errors in the \pos tagger and/or the parser. Errors resulting from \pos errors are very minimal, and so we do not discuss them further here.  Some of the parser errors are of the type that parsers typically have problems with - attachment and coordination.  For example, two of the recall errors for {\tt verb-not} have to do with  {\tt not}  appearing not as the sister to the verb, but rather as the first word in the subordinate clause immediately following.  Such errors are unlikely to completely avoided.

However, we focus here on a few errors are of a different type and that also account for the main source of  errors - the recall errors for {\tt do-subj} and {\tt verb-subject}.  These errors all have the feature that the parser is creating structures that are unattested in the training data, having to do with the interaction of the {\tt CP} and {\tt IP} levels.

The top pair of trees in Figure \ref{ds-recall} show a case in which the tree output by the parser matches exactly the gold tree, including the function tags, except for the root node, for which it is has a {\tt IP-MAT} instead of
a {\tt CP-QUE-MAT}, and therefore does not match 
{\tt do-subj} query. The structure {\tt (IP-MAT (IP-SUB ...DOD...))}  
does not occur in the training data, although it is generated by the parser in this case. 

The bottom pair of trees in Figure \ref{ds-recall} illustrates another recall error.  In this case, while the parser of course does not get the co-indexing and empty category, as discussed in Section \ref{sec:parser:ftags}, there is a serious structural error, in that it does not make the {\tt IP-SUB} tree, which causes
it to not match the {\tt do-subj} query.  The structure {\tt (CP-QUE-MAT ... DOP ...)} in the parser output does
not occur in the training data. 

Turning now to {\tt verb-subject} recall errors, 
Figure \ref{vs-recall} shows a gold tree in which the parenthetical {\tt thynke you} is the {\tt verb-subject}.  The parser output
does not get the parenthetical structure. While a parenthetical 
clause is not always easy to parse correctly, especially without surrounding
parentheses, it is noteworthy that the resulting {\tt IP-SUB} has two {\tt NP-SBJ}s, something which certainly does not exist in the training data.  This example is in fact the only case of the {\tt non-inverted}, because the first {\tt NP-SBJ} precedes the {\tt VBP}.

These examples suggest that a fruitful area of parser analysis would be to create searches for various ``impossible'' structures and so how many exist in the parser output.

\subsection{Partial and full parses}
\label{sec:queries:fullparses}
As discussed in Section \ref{sec:parser:ftags}, while the parser is able to output ten of the function tags, it does not generate the others, and it is not able to output empty categories, with the co-indexing with overt elements when appropriate. The limited output has been 
appropriate for this first stage of the work, but this is an area for future work.  We have only discussed a few of the relevant searches, and some of the others
we have not yet tried rely on some of the function tags that the parser cannot currently produce, and which are quite different from function tags considered 
in parsing work. For example,  one search relies on the {\tt TMC} tag, for ``tough movement complement''.  Likewise, although the lack of empty categories and co-indexing has not caused serious issues for the current
queries, as the range of queries is expanded to include some that test for syntactic movement, this will increase in importance as an issue.

In short, further work will require a movement toward ``fully parsing'' the \ppceme and \eebo, in the sense of \citet{gabbard-etal-2006-fully}, recovering more of the function
tags and outputting the empty categories and co-indexing.

\section{Conclusion and Future Work}
\label{sec:conclusion}

We have described the first stages of an overall project of using modern NLP techniques to automatically annotate large amounts of text for linguistic search, with promising results
for the queries discussed.   Future work will proceed in the following areas: \\
{\bf Evaluation} We will expand the parsing and query evaluation to use a 10-fold split of the \ppceme, to obtain a more reliable measure of the results. We will also implement the suggestion at the end of Section \ref{sec:queries:errors} to create searches for various ``impossible'' structures.  \\
{\bf Parsing} We will use this improved evaluation metric to experiment with different parsing models and improved embeddings \cite{devlin-etal-2019-bert}.  We will include more of the \ppceme function tags in the parser model, as well as aim to recover empty categories with their co-indexing.  As mentioned at the end of Section \ref{sec:queries:fullparses}, this will be an issue for queries beyond the ones discussed in this paper.  Work in this area will be driven by the needs of the linguistic queries, in that we will focus on the function tags and empty categories needed for the particular queries, rather than necessarily tackling the entire problem at once. \\
{\bf EEBO} Of course the parsing and query search  will be carried out on \eebo as well.  We also plan with our linguistic collaborators to do a certain amount of gold-standard annotation on a sample of \eebo, for a direct evaluation of the NLP infrastructure on the target corpus.

\appendix

\section{Corpus preparation}
\label{app:corpus}

\subsection{\ppceme}
\label{app:corpus:ppceme}

In addition to the changes described in section \ref{sec:eebo}, we removed the metadata included in {\tt CODE}, {\tt META}, and {\tt REF} elements. For 267 trees a leaf of the tree was inside a {\tt CODE} element, and since removing this information resulted in an ill-formed tree, we did not include these trees. 576 trees were rooted in {\tt META} (usually stage directions for a play) and we removed those trees. We also removed 9 trees with a {\tt BREAK} element.

In addition, before doing the above, we changed all instances of \verb+(CODE <paren>)+ and \verb+CODE <$$paren>)+ to \verb+(OPAREN -LRB-)+ and \verb+(CPAREN -RRB-)+, respectively, so they would not be included in the {\tt CODE} removal since we wanted to retain the parentheses. 

We note that our counts for number of words and sentences differ slightly from  \citet{yang-eisenstein-2016-part}, which is probably related to small differences of preparation that we aim to resolve in the future.

\subsubsection{Partitioning}
\label{app:corpus:ppcme:part}
The dev section consists of the 16 files beginning with {\tt l}, which cover the time span $1539-1696$. The test section consits of the 31 files beginning with {\tt e}, which cover the time span $1501-1690$.  The rest of the corpus covers the time span $1502-1719$.

\subsubsection{Part-of-speech tags}
\label{app:corpus:ppceme:pos}
In addition to the changes described in the main text (the complex tags and removing the numbers from tags), we also changed the tag {\tt MD0} to {\tt MD}. {\tt MD0} is an untensed modal, as in {\tt he will can} or {\tt to can do something}. There are only four cases, as this is an option that had mostly died out by the time of Early Modern English.

\subsection{EEBO}
\label{app:corpus:eebo}
In addition to the normalizations discussed in section \ref{sec:eebo}, following \cite{ecay}, we removed information under the {\tt NOTE}, {\tt SPEAKER}, {\tt L},and {\tt GAP} tags.  The {\tt L} is ``lyrical'' text - e.g, song lyrics - which was not appropriate for the searches of linguistic interest. In future work we will likely revise this to keep this text but with some meta-tags to indicate its origin.

\subsubsection{Preprocessing for \elmo}
\label{app:corpus:eebo:training}
The version of \eebo used for \elmo training is slightly different from that described in section \ref{sec:eebo}. While the extraction, normalization, and tokenization steps are identical, we did not perform any sentence segmentation prior to \elmo training. We did also exclude all text lines with one of the characters that occurred fewer than 200 times in the corpus.  This eliminated 4139 lines, with 9,341,966 remaining (consisting of 1,168,749,620 tokens) for training.

\label{app:training}
\section{Training details}
\subsection{ELMo embeddings}
\eebo \elmo embeddings were trained using TensorFlow maintained distributed by AllenNLP at \url{https://github.com/allenai/bilm-tf} using the default model configuration. For the contemporary \elmo embeddings (referred to as {\tt original} in Table \ref{tab:pos}, we used the {\tt elmo\_2x4096\_512\_2048cnn\_2xhighway} model from the AllenNLP website, which was trained on the 1B Word Benchmark.

\subsection{POS tagger}
\label{app:training:pos}
 The \pos tagger was trained using AllenNLP v0.8.5. For the learned word embeddings we used 50 dimensions. For the character based representations we used 128 convolutional filters and 16 dimensional character embeddings. All LSTM layers used a hidden dimension size of 200. We trained using a batch size of 6,000 tokens and the Adam optimizer for up to 75 epochs using early stopping. For further details regarding model initialization and training, we refer readers to the AllenNLP constituency parsing configuration file {\tt \_config.ner\_elmo.jsonnet}.

\subsection{Parser}
\label{app:training:parser}
\subsubsection{Model parameters and training}
Both the LSTM layers from the decoder and the feedforward layer that processes the spans used 250 hidden units. Versions of the model using learned word embeddings and character based CNN representations used the same parameters as for the \pos tagger  --  50 and 128, respectively. 

Training was performed using all sentences from the \ppceme training section of length $\leq$ 300, which resulted in the exclusion of 65 sentences (out of 85,398). The parser was trained using a batch size of 500 tokens and the Adadelta optimizer for up 100 epochs using early stopping.

As with the \pos tagger, we used \allennlp v0.8.5. For further details regarding model initialization and training, we refer readers to the AllenNLP constituency parsing configuration file {\tt constituency\_parser\_elmo.jsonnet}.

\begin{table*}[ht]
\begin{minipage}[b]{.68\textwidth}
    \centering
    {\small
    \begin{tabular}{|l|c|c|c|c|c|c|}   \cline{2-7}
    \multicolumn{1}{c}{ } & 
    \multicolumn{3}{|c|}{{\bf With Function Tags}}  & 
    \multicolumn{3}{|c|}{{\bf Without Function Tags}} \\
    \hline
    {\bf Embeddings}     &  {\bf Recall} &  {\bf Prec} &  {\bf F1} &  {\bf Recall} &  {\bf Prec} &  {\bf F1}  \\
    \hline
    (1) \eebo            &         87.95 &       88.64 &     88.30 &         88.39 &       88.04 &     88.21  \\
    \hline
    (2) original         &         84.91 &       85.32 &     85.11 &         84.57 &       84.85 &     84.71  \\
    \hline
    (3) none, tokens    &         80.48 &       84.01 &     82.20 &         81.05 &       83.39 &     82.20  \\
    \hline
    (4) \eebo, tokens    &         88.02 &       88.26 &     88.14 &         88.08 &       88.44 &     88.26 \\
    \hline
    (5) Berkeley         &         79.98 &       74.95 &     77.37 &         80.08 &       75.38 &     77.66 \\
    \hline
    \end{tabular}}
    \caption{Evalb precision, recall, and F1 for the \ppceme dev section, with and without function tags. See Table \ref{tab:results:parsing} for explanation of first column.}
    \label{app:results:parsing}
    \end{minipage}
    \begin{minipage}[b]{.32\textwidth}
    \centering
    {\small 
    \begin{tabular}{|l|r|r|}  \hline
    {\bf Tag} &  {\bf \# gold} &  {\bf F1} \\
    \hline
    SBJ       &           6959 &     98.02 \\
    \hline
    MAT       &           3646 &     98.29 \\
    \hline
    SUB       &           4180 &     99.15 \\
    \hline
    ACC       &           3803 &     96.17 \\
    \hline
    INF       &           1715 &     98.45 \\
    \hline
    PRN       &            830 &     89.20 \\
    \hline
    QUE       &            259 &     94.46 \\
    \hline
    IMP       &            129 &     94.07 \\
    \hline
    DTV       &            384 &     91.56 \\
    \hline
    VOC       &            126 &     93.13 \\
    \hline
    \end{tabular}}
    \caption{Function tag results for \ppceme test set.}
    \label{app:results:ftags}
    \end{minipage}
\end{table*}

\subsubsection{Function Tags}
\label{app:training:parser:ftags}

See \url{https://www.ling.upenn.edu/hist-corpora/annotation/labels.htm} for more discussion of the tags. We are excluding five tags - {\tt YYY, ELAB, XXX, TPC, TAG} - that only occur a total of 25 times altogether.

\section{Parser results on \ppceme test section}
\label{app:parser:ppceme:results}

In Tables \ref{app:results:parsing} and \ref{app:results:ftags} we show the results for the parsing and function tag evaluation on the test section, analogous to Tables \ref{tab:results:parsing} and \ref{tab:results:ftags} in Section \ref{sec:parser:results}.   While the parsing results follow the same general pattern as with the dev section, the numbers are a bit lower.  We assume that this is due to random differences between the dev and test sections, but it does point out the importance of using a 10-fold evaluation, as mentioned in the conclusion.   The function tag evaluation shows a bit more difference, with the order by frequency of some of tags changing. However, we kept the tags in the same order as in Table \ref{tab:results:ftags}.   The scores on some of the less-frequently occurring function tags (e.g., {\tt IMP} and {\tt VOC}) drop a bit compared to the dev section, again showing the importance of doing a 10-fold evaluation.

\clearpage

\bibliography{ppcbib}
\bibliographystyle{acl_natbib}

\end{document}